\documentclass[sigconf,usenames,dvipsnames]{acmart}
\copyrightyear{2021} 
\acmYear{2021} 
\setcopyright{acmcopyright}\acmConference[KDD '21]{Proceedings of the 27th ACM SIGKDD Conference on Knowledge Discovery and Data Mining}{August 14--18, 2021}{Virtual Event, Singapore}
\acmBooktitle{Proceedings of the 27th ACM SIGKDD Conference on Knowledge Discovery and Data Mining (KDD '21), August 14--18, 2021, Virtual Event, Singapore}
\acmPrice{15.00}
\acmDOI{10.1145/3447548.3467215}
\acmISBN{978-1-4503-8332-5/21/08}

\usepackage{wrapfig}
\usepackage{lipsum}
\usepackage{subfigure}
\usepackage{graphicx} 
\usepackage{float}
\usepackage{amsmath}
\usepackage{amsfonts}
\usepackage{color}
\usepackage{comment}
\usepackage{cancel}
\usepackage{soul}
\usepackage[switch]{lineno}  %
\usepackage{newtxmath}
\usepackage[e]{esvect}
\usepackage{booktabs}
\usepackage{footnote}
\usepackage{array}
\usepackage{bm}
\usepackage{multirow}
\usepackage{caption}
\usepackage{sidecap}
\usepackage{algorithm,algorithmic}

\newcommand{\ahm}{Adaptive Hybrid Masked Model}
\newcommand{\sahm}{AHM}
\newcommand{\np}{Cross Entity Alignment}
\newcommand{\snp}{CEA}

\newcommand{\ec}{e-commerce}
\newcommand{\tb}{\textbf}
\newcommand{\ti}{\textit}
\newcommand{\mc}{\mathcal}

\newcommand{\bs}{\boldsymbol}

\def\rvu{{\mathbf{i}}}

\def\rvu{{\mathbf{u}}}
\def\rvv{{\mathbf{v}}}

\def\ervu{{\textnormal{u}}}
\def\ervv{{\textnormal{v}}}

\def\rmP{{\mathbf{P}}}

\DeclareMathAlphabet{\mathsfit}{\encodingdefault}{\sfdefault}{m}{sl}
\SetMathAlphabet{\mathsfit}{bold}{\encodingdefault}{\sfdefault}{bx}{n}

\def\sR{{\mathbb{R}}}

\def\sX{{\mathbb{X}}}
\def\sY{{\mathbb{Y}}}

\DeclareMathOperator*{\argmin}{arg\,min}

\newcommand{\Tmat}{{\bf T}}

\newcommand{\xv}{{\boldsymbol x}}
\newcommand{\yv}{{\boldsymbol y}}

\newcommand{\gammav}{{\boldsymbol \gamma}}

\newcommand{\muv}{{\boldsymbol \mu}}
\newcommand{\nuv}{{\boldsymbol \nu}}

\newcommand{\ie}{\textit{i.e.}}

\definecolor{darkblue}{rgb}{0.0, 0.0, 0.55}
\definecolor{darkred}{rgb}{0.55, 0.0, 0.0}

\title{Domain-oriented Language Modeling with Adaptive Hybrid Masking and Optimal Transport Alignment}

\author{Denghui Zhang$^{1}$, Zixuan Yuan$^{1}$, Yanchi Liu$^{2*}$, Hao liu$^{4}$,\\ Fuzhen Zhuang$^{3}$, Hui Xiong$^{1*}$, Haifeng Chen$^2$
}\thanks{* Corresponding authors. This work was partially supported by the National Science Foundation through awards III-2006387, IIS 1814510, and IIS-2040799.}

\affiliation{%
  \institution{$^1$Rutgers, The State University of New Jersey, 
  $^2$NEC Labs America, USA\\
        $^3$Institute of Artificial Intelligence, School of Computer Science, Beihang University, China \\
        $^4$The Hong Kong University of Science and Technology, China\\
    }
}

\begin{document}
\settopmatter{printacmref=true} 
\fancyhead{} 

\begin{abstract}
Motivated by the success of pre-trained language models such as BERT in a broad range of natural language processing (NLP) tasks, recent research efforts have been made for adapting these models for different application domains. Along this line, existing domain-oriented models have primarily followed the vanilla BERT architecture and have a straightforward use of the domain corpus. 
However, domain-oriented tasks usually require accurate understanding of domain phrases, and such \ti{fine-grained {phrase-level} knowledge} is hard to be captured by existing pre-training scheme. 
Also, the word co-occurrences guided semantic learning of pre-training models can be largely augmented by \ti{entity-level association knowledge}.
But meanwhile, by doing so there is a risk of introducing noise due to the lack of groundtruth word-level alignment. 
To address the above 
issues, 
we provide a generalized domain-oriented approach, which leverages auxiliary domain knowledge to improve the existing pre-training framework from two aspects.
First, to preserve phrase knowledge effectively, we build a domain phrase pool as auxiliary training tool, meanwhile we introduce \ahm~to incorporate such knowledge. 
It integrates two learning modes, word learning and phrase learning, and allows them to switch between each other.
Second, we introduce \np~to leverage entity association as weak supervision to augment the semantic learning of pre-trained models.
To alleviate the potential noise in this process, we introduce an interpretable \ti{Optimal Transport based approach} to guide alignment learning.
Experiments on four domain-oriented tasks demonstrate the superiority of our framework.
\end{abstract}

\ccsdesc[500]{Computing methodologies~Natural language processing}
\keywords{Domain language modeling, pre-training, masked language model, optimal transport.}

\maketitle

\section{Introduction}
\begin{figure}[!t]
\setlength{\belowcaptionskip}{-6pt}
\centering
\includegraphics[width=0.478\textwidth]{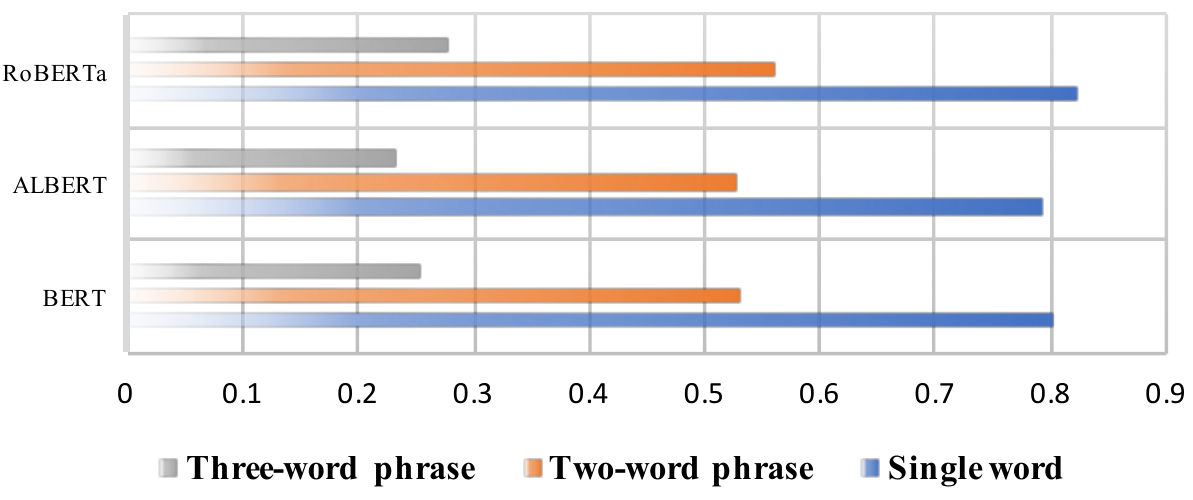}
\vspace{-0.5cm}
\caption{The single-word and phrase reconstruction accuracy of several existing language pre-training models.}
\label{intro-mlm}
\vspace{-0.2cm}
\end{figure}
\begin{table}[t]
\centering
\caption{An example of review aspect extraction, where correct answers (marked in color) are usually phrases. }
\vspace{-0.3cm}
\includegraphics[width=0.47\textwidth]{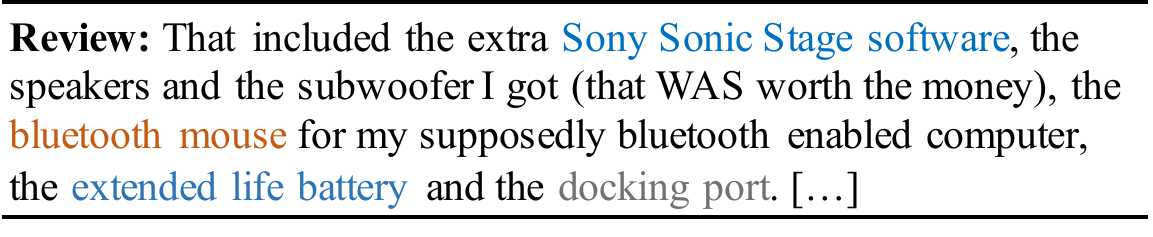}
\label{intro}
\vspace{-0.8cm}
\end{table}
Recent years have witnessed the great success of pre-trained language models (PLMs), such as BERT~\cite{devlin2019bert}, in a broad range of natural language processing (NLP) tasks.
Moreover, several domain-oriented PLMs have been proposed to adapt to specific domains \cite{huang2019clinicalbert, gu2020domain, chalkidis2020legal}.  
For instance, BioBERT \cite{lee2020biobert} and SciBERT \cite{beltagy2019scibert} are pre-trained leveraging large-scale domain-specific corpora for biomedical and scientific domain tasks respectively.
However,
in the above models,
the same pre-training scheme as BERT is reused straightforwardly, while insightful domain characteristics are largely overlooked.
To this end, we raise a natural question: 
\ti{for domain language pre-training, can we go further beyond the strategy of vanilla BERT + domain corpus by leveraging domain characteristics?}
In this paper, we explore this question under \ec~domain and present promising approaches that can also be generalized to other domains when auxiliary knowledge is available. 

We first discuss the characteristics of domain-oriented tasks, 
and the limitations of current pre-training approaches, 
then present two major improving strategies, 
corresponding to leveraging two types of auxiliary domain knowledge smartly.
On the one hand, understanding a great variety of  \ti{domain phrases} is critical to domain-oriented tasks.
As shown in Table \ref{intro}, the review aspect extraction task, widely used in the \ec~domain, requires language models to understand domain phrases to extract the correct answers.
However, such \ti{phrase-level} domain knowledge is hard to be captured by Masked Language Model (MLM) \cite{devlin2019bert} (i.e., the self-supervised task employed in most language pre-training models).
Figure \ref{intro-mlm} depicts the language reconstruction performance of three existing language pre-training models on a public \ec~corpus. 
As can be seen,
the reconstruction accuracy drops drastically when the prediction length is increased from single word to multi-word phrase.
We attribute this to the fact that MLM is a \ti{word-oriented} task, i.e., it only reconstructs randomly masked words from the incomplete input however does not explicitly encourage any perception ability for domain phrases.
Although later works \cite{joshi2020spanbert,sun2020ernie} propose to mask phrases instead of words in MLM to enable BERT for phrase perception, 
they have two major drawbacks:
(\romannumeral1) 
\ti{Overgeneralized phrase selection},
they use chunking \cite{sun2019ernie} to randomly select phrases to mask, without considering the quality of phrases and the relatedness to specific domains.
(\romannumeral2)
\ti{Discard of word masking}, 
word masking helps to acquire word-level semantics essential for phrase learning, hence should be preserved in pre-training.

On the other hand,
pre-trained language models are limited by corpus-level statistics such as co-occurrence, which can be mitigated by auxiliary domain knowledge.
For instance, to learn that \texttt{Android} and \texttt{iOS} are semantically related, a large number of co-occurrences in similar contexts are required in the pre-training data.
For domain-oriented learning, this can be mitigated by auxiliary knowledge, i.e., 
\ti{entity association}.
As shown in Table \ref{intro2}, 
when leveraging the ``substitutable'' association to pair the description texts of two product entities, \texttt{Samsung galaxy} and \texttt{iPhone}, 
we can augment the co-occurrence of some words/phrases (e.g, \ti{5G network} vs \ti{4G signal}; \ti{Android} vs \ti{iOS})
by learning the alignments of similar words across entities.
However, the above intuition is challenging to fulfill in practice as it constitutes a \ti{weakly supervised learning task}.
In other words, only weak-supervision signals (i.e., entity-level alignments) are available, 
while the word-level groundtruth alignments across entities are hard to obtain.
Hence, the aligning problem needs a \ti{robust} learning algorithm to overcome the potential noises under the weak supervision.
Moreover, the algorithm should also offer decent \ti{interpretability} over the alignment for the ease of understanding and validation.

Based on the above insights, 
we propose an enhanced domain-oriented framework for language pre-training.
Our framework takes the mentioned domain characteristics into consideration,
and introduces two approaches to tackle the challenges.
First, to enable language pre-training with the perception ability for domain phrases, 
we propose an advanced alternative for Masked Language Model, namely, \ahm~(\sahm).
In contrast to MLM only masking and reconstructing single words, 
\sahm~introduces a new sampling scheme for masking quality phrases with the guidance of an external domain phrase pool, and meanwhile, 
a novel \ti{phrase completeness regularization term} is proposed for sophisticated phrase reconstruction.
Furthermore,
since both word-level and phrase-level semantics are critical to language modeling, 
we unify the word and phrase learning modes via a loss-based parameter. 
It allows the adaptive switching between each other, ensuring a \ti{smooth and progressive} learning process resembling the human cognition of language.
\begin{table}[t]
\centering
\caption{An example of relational text in the e-commerce domain, where product descriptions are connected by the ``substitutable'' product association.}
\vspace{-0.2cm}
\includegraphics[width=0.47\textwidth]{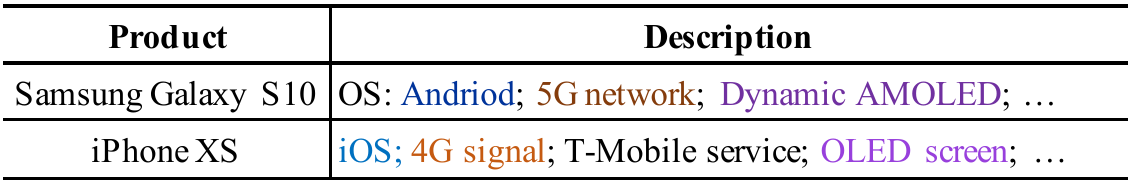}
\label{intro2}
\vspace{-0.7cm}
\end{table}
Second, to exploit the rich co-occurrence signals hidden in entity associations, 
we formulate a new pre-training task, namely, \np~ (\snp).
Specifically,
CEA aims to learn the word-level alignment matrix of entity association based text pair (e.g., description pair) with only weak supervision, i.e., only knowing two entities are related but no word-level groundtruth alignments available.
Moreover, we propose an alignment learning scheme leveraging Optimal Transport (OT) to train this task in a weakly-supervised fashion.
At each round, the OT objective helps to find the pseudo optimal matching of similar words (or phrases) and returns a \ti{sparse transport plan}, which reveals robust and interpretable alignments.
The language model is further optimized with the guidance of the transport plan to minimize the Wasserstein Distance of the aligned entity contents, enabling the model to learn fine-grained semantic correlations.

To validate the effectiveness of the proposed approach, we conduct extensive experiments in the \ec~domain to compare our pre-training framework with state-of-the-art baselines.
Specifically, we employ the pre-training corpus created from publicly available resources and fine-tune on four downstream tasks, i.e., Review-based Question Answering (RQA), Aspect Extraction (AE), Aspect Sentiment Classification (ASC), and Product Title Categorization (PTC).
Quantitative results show that our method significantly outperforms BERT and other variants on all the tasks.
Additionally, the visualization of OT-based approach reveals feasible alignment results despite the weak supervision, 
meanwhile, presenting convincing interpretability as the alignment vector is enforced to be sparse. 
Lastly, while we demonstrate the effectiveness of our approach in the \ec~domain, the ideas of the framework can be generalized to broader domains 
since the aforementioned auxiliary knowledge is free of annotation cost. 
The domain phrase pool can be constructed from domain corpus.  
Entity association is broad and general, which is easy to obtain in main domains.


\section{Related Work}
\noindent\tb{Pre-trained Language Models.}
Recently, the emergence of pre-trained language models (PLMs) \cite{devlin2019bert,peters2018deep,radford2018improving} has brought natural language processing to a new era.
Compared with traditional word embedding models \cite{mikolov2013distributed}, PLMs learn to represent words based on the entire input context to tackle polysemy, hence captures semantics more accurately.
Following PLMs, many endeavors have been made for further optimization in terms of both architecture and training scheme \cite{liu2019roberta,brown2020language,sun2020ernie,liang2020bond}.
Along this line, SpanBERT \cite{joshi2020spanbert} proposes to reconstruct randomly masked spans instead of single words. 
However, the span consists of random continuous words and may not form phrases, thus fails to capture phrase-level knowledge effectively.
ERNIE \cite{sun2019ernie} integrates phrase-level masking and entity-level masking into BERT, which is closely related to our masking scheme.
Differing from their work simply using chunking to get general phrases, we build high-quality domain phrase pool to assist learning domain-oriented phrase knowledge.
Also, we propose a novel phrase regularization term over the reconstruction loss to encourage complete phrase learning.
Moreover, we combine word and phrase learning cohesively according to their optimizing progress, 
achieving better performance than each single mode.

\noindent{\tb{Domain-oriented PLMs.}}
To adapt PLMs to specific domains, several domain-oriented BERTs such as
BioBERT \cite{lee2020biobert}, SciBERT \cite{beltagy2019scibert}, 
and TweetBERT \cite{qudar2020tweetbert}, 
have been proposed recently.
BERT-PT \cite{xu2019bert} proposes to post-train BERT on a review corpus and obtains better performance on the task of review reading comprehension.
\citeauthor{gururangan2020don} \cite{gururangan2020don} proposes an approach for post-training BERT on domain corpus as well as task corpus to obtain more performance gains on domain-specific tasks.
DomBERT \cite{xu2020dombert} proposes to select data from a mixed multi-domain corpus for the target domain, improving the diversity of domain language learning.
More work along this line can be referred to \cite{rietzler2020adapt,ma2019domain}.
Similarly, 
incorporating domain knowledge has shown effectiveness in broader areas \cite{yuan2020spatio,zhang2017efficient,sun2021market,zhang2019job2vec,Yuan_Liu_Hu_Zhang_Xiong_2021,li2018link} such as representation learning.
The above solutions have primarily leveraged domain corpus for pre-training in a straightforward way, without considering insightful domain characteristics and domain knowledge such as domain phrase and entity association.
Our work is the first leveraging auxiliary domain knowledge to enhance domain-oriented pre-training.
\section{Preliminaries}

In this section, we give a brief introduction to two essential concepts that are related to our work, namely, Masked Language Model and Optimal Transport.


\noindent{\tb{Masked Language Model}}.
Masked Language Model (MLM) \cite{devlin2019bert} refers to the self-supervised pre-training task that have been applied in pre-trained language models (e.g., BERT, RoBERTa, etc.). 
It is considered as a fill-in-the-blank task, i.e., 
given an input sequence partially masked (15\% tokens), 
it aims to predict those masked words using the embeddings generated by the language model:
\begin{equation}
p\big(\footnotesize{ X_m\big|X_{\backslash M}}\big)=\displaystyle\frac{\exp\left(\bs{W_m^\top}\big[\mc{F}\big(X_{\backslash M};\theta\big)\big]_m\right)}{\displaystyle\sum\limits_{k\in \mc{V}}\exp\left(\bs{W_k^\top}\big[\mc{F}\big(X_{\backslash M};\theta\big)\big]_m\right)},
\end{equation}
where $\mc{F}(;\theta)$ denotes the Transformer based language model.
$X$ is the full input sequence,
$M$ denotes the indices of all masked tokens in $X$,
$X_m$ indicates one of the tokens in $M$, 
$\backslash$ is set minus.
$[\mc{F}\big(X_{\backslash M};\theta\big)\big]_m$ denotes the output vector corresponding to the masked token $X_m$ and $W^\top$ denotes the softmax matrix with the same number of entries as the vocabulary $\mc{V}$.

Maximizing $p\big(\footnotesize{ X_m\big|X_{\backslash M}}\big)$ enforces $\mc{F}(;\theta)$ to infer the meaning of masked words from their surroundings, in other words, preserving contextual semantics.

\noindent\tb{Optimal Transport and Wasserstein Distance}.
Optimal Transport (OT) studies 
the problem of
transforming one probability distribution into another one (e.g., one group of embeddings to another) with the lowest cost.
When considering the ``cost'' as distance, a commonly used  distance metric for OT is Wasserstein Distance (WD) \cite{villani2008optimal}.
Formal definition is as follows\cite{chen2020graph}:
\begin{definition}
	\label{def:wd}
	Let $\muv \in \rmP(\sX), \nuv \in \rmP(\sY)$ denote two probability distributions, formulated as $\muv = \sum_{i=1}^m \ervu_i \delta_{\xv_i}$ and $\nuv = \sum_{j=1}^n \ervv_j \delta_{\yv_j}$, with $\delta_{\xv}$ as the Dirac function centered on $\xv$. $\Gamma (\muv,\nuv)$ denotes all the couplings (joint distributions) of $\muv$ and $\nuv$, with marginals $\muv(\xv)$ and $\nuv(\yv)$.
	The optimal Wasserstein Distance between the two distributions $\muv,\nuv$ is defined as:
	\begin{align}\label{eq:wd}
		\mathcal{D}_{w}(\muv,\nuv) = & \inf_{\gammav\in\Gamma (\muv,\nuv)}\mathbb{E}_{(\xv,\yv)\sim\gammav}\,\, [c(\xv,\yv)] \nonumber\\
		=\min_{\Tmat\in \Gamma (\rvu,\rvv)}\left \langle \tb{T}, \tb{C} \right \rangle&=\min_{\Tmat\in \Gamma (\rvu,\rvv)}\sum_{i=1}^m \sum_{j=1}^n \Tmat_{ij} \cdot c(\xv_i,\yv_j)\,,
	\end{align}
	where $\Gamma (\rvu,\rvv) = \{ \Tmat \in \sR_+^{m\times n} | \Tmat\mathbf{1}_n=\rvu, \Tmat^\top\mathbf{1}_m=\rvv \} $, $\mathbf{1}_m$ denotes an $m$-dimensional all-one vector,
	the weight vectors $\rvu=\{\ervu_i\}_{i=1}^m \in \Delta_m$ and $\rvv=\{\ervv_i\}_{i=1}^n \in \Delta_n$ belong to the $m$- and $n$-dimensional simplex, respectively (\ie, $\sum_{i=1}^m \ervu_i = \sum_{j=1}^n \ervv_j = 1$).
	And $c(\xv_i,\yv_j)$ is the cost function evaluating the distance between $\xv_i$ and $\yv_j$ (samples of the two distributions). 
	Computing the optimal distance (1st line) is equivalent to solving the network-flow problem (2nd line) \cite{luise2018differential}.
	The calculated matrix $\Tmat$ denotes the ``transport plan'', 
    where each element $\Tmat_{ij}$ represents the amount of mass shifted from $\ervu_i$ to $\ervv_j$.
    We propose an Optimal Transport based approach for the cross entity alignment problem in Section 4.2.
\end{definition}

\section{Methodology}

\begin{figure}[t]
\setlength{\belowcaptionskip}{-6pt}
\centering
\includegraphics[width=0.47\textwidth]{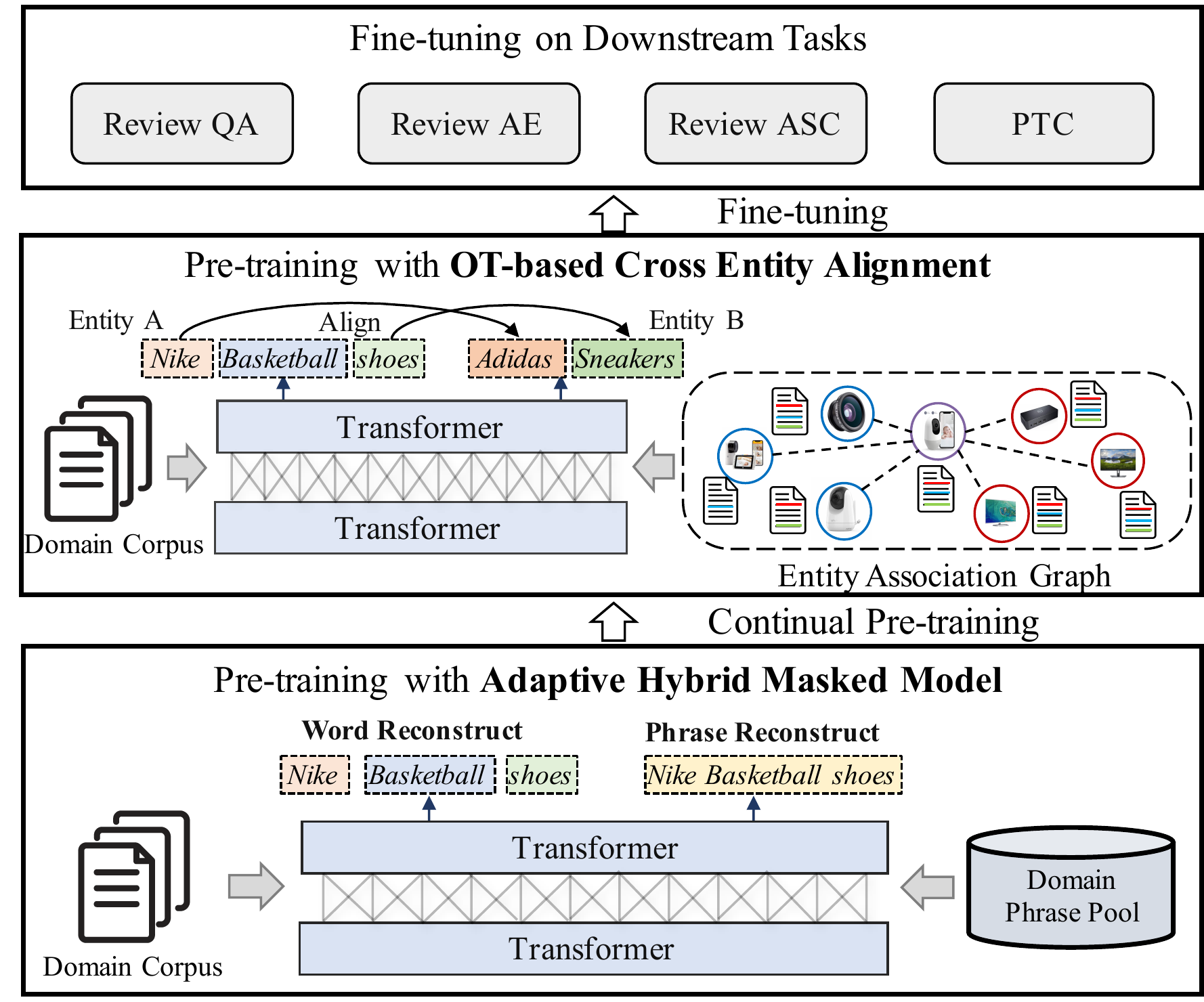}
\vspace{-0.2cm}
\caption{ Framework overview.
}
\label{ov}
\vspace{-0.3cm}
\end{figure}
\begin{figure*}[!t]
\setlength{\belowcaptionskip}{-6pt}
\centering
\includegraphics[width=0.975\textwidth]{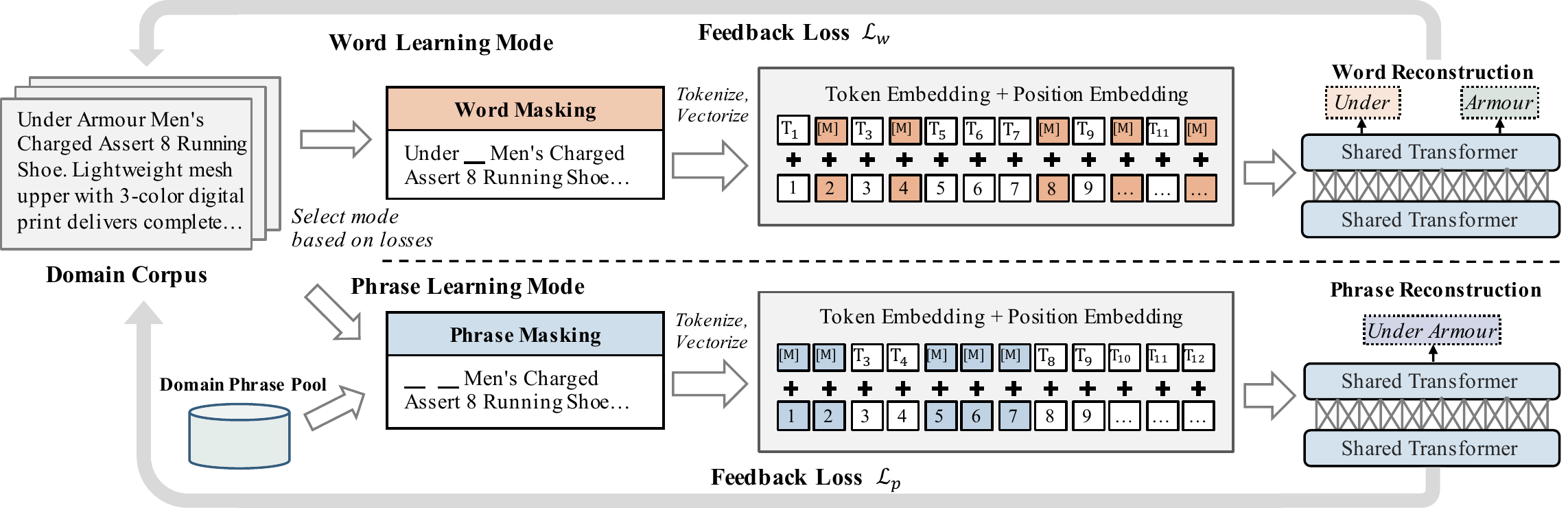}
\vspace{-0.1cm}
\caption{
Illustration of \ahm. 
Based on the feedback losses, it adaptively switches between two learning modes,
enabling the language model to learn word-level and phrase-level knowledge simultaneously.
}
\label{ahm}
\vspace{-0.2cm}
\end{figure*}

In this section,
we provide an in-depth introduction to our enhanced framework for domain-oriented language pre-training.
Figure \ref{ov} presents an overview of the framework,
consisting of two major improvements, i.e., \ahm~(\sahm) to replace MLM and a new weakly-supervised pre-training task, 
OT-based \np~(\snp).
The former leverages a domain corpus and a domain phrase pool to learn both word-level and phrase-level semantics, the latter utilizes the same corpus and an entity association graph to obtain text pairs for augmenting domain semantic learning.
Moreover, we employ continual multi-task pre-training \cite{sun2020ernie} to jointly train \sahm~and \snp.
Lastly, the model is fine-tuned to be deployed in domain-oriented applications.

\subsection{\ahm}
In order to enhance the phrase perception ability of language model while meantime preserving its original word perception ability, we introduce a new masked language model, namely, \ahm~(\sahm).
Specifically,
we set two learning modes in \sahm, i.e., \ti{word learning} and \ti{phrase learning}, 
which in a nutshell, \ti{masks then reconstructs} word units and phrase units, respectively.
Moreover, we combine the two learning modes by adaptively switching between them, enabling the model to capture the word-level and phrase-level semantics \ti{simultaneously} and \ti{progressively}. 
Figure \ref{ahm} provides an illustration of the model.

\subsubsection{\tb{Word Learning Mode}}
In this mode, given an input sequence $X^{t}$ ($t$ denotes the $t^{th}$ iteration), 
we first randomly sample words from $X^{t}$ iteratively until the selected words constitute 15\% of all tokens.
Then we replace them with: (1) the \texttt{[MASK]} token 80\% of the time, (2) a random token 10\% of the time, (3) the original token 10\% of the time.
Next, we predict all the masked/perturbed tokens 
by feeding their embeddings of the language model to a shared softmax layer. 
Equivalently, we optimize the log-likelihood function below:
\begin{equation}
\mc{L}_{w}=-\log\prod_{m\in{\mc{W}^t}}p\big({ X^{t}_m\big|X^{t}_{\backslash \mc{W}^t}}\big),
\end{equation}
where $\mc{W}^t$ denotes the indices of all the masked/perturbed tokens in $X^{t}$.
$X^{t}_m$ and $X^{t}_{\backslash \mc{W}^t}$ denotes the $m^{th}$ masked token and perturbed input, respectively.
$p\big(X^{t}_m\big|X^{t}_{\backslash \mc{W}^t}\big)$ follows the definition in Eq.(1).
This mode resembles the original masking scheme in MLM except that we only mask whole words. 
It helps to learn preliminary word-level semantics, which is not only the basis of language understanding but also essential for phrase learning. 
\subsubsection{\tb{Phrase Learning Mode}}
In the phrase learning mode, we randomly mask consecutive tokens that constitutes \ti{quality domain phrases} and train the language model to reconstruct them.
First, given an input sequence $X^t$ and a domain phrase pool $\mc{P}_{D}$ (comprising high-quality phrases and their quality scores)\footnote{In this paper, we leverage AutoPhrase~\cite{shang2018automated} to obtain domain phrase pool.}, following Algorithm \ref{al:dp}, we detect domain phrases and sample to obtain $15\%$ tokens.
Then similar to the word mode, we replace the selected tokens with \texttt{[MASK]} token 80\% of the time, a random token and the original token 10\% of the time respectively.
Next, we optimize the following loss function to reconstruct the masked phrases:
\begin{equation}
\small
\mc{L}_{p}=-\Big(\log\prod_{m\in{\mc{P}^t}}p\big({ X^{t}_m\big|X^{t}_{\backslash \mc{P}^t}}\big)+
\underbrace{\log\prod_{P\in{\hat{\mc{P}}^t}}r\big({ X^{t}_P\big|X^{t}_{\backslash \hat{\mc{P}}^t}}\big)}\limits_{\ti{completeness regularization}}\Big),
\end{equation}
\begin{equation}
\small
r\big({ X^{t}_P\big|X^{t}_{\backslash \hat{\mc{P}}^t}}\big)=
\displaystyle\frac{\exp\left(\bs{{C}_P^\top}\tb{Avg}\big(\big[\mc{F}\big(X^{t}_{\backslash \hat{\mc{P}}^t};\theta\big)\big]_{P}\big)\right)}{\displaystyle\sum\limits_{k\in \mc{V}_p}\exp\left(\bs{{C}_k^\top}\tb{Avg}\big(\big[\mc{F}\big(X^{t}_{\backslash \hat{\mc{P}}^t};\theta\big)\big]_{P}\big)\right)},
\end{equation}
where the first term is defined the same way as Eq.(1) and (3) except that
$\mc{P}^t$ denotes indices of all the masked tokens obtained via Algorithm \ref{al:dp}.
With the first term, we \ti{reconstruct masked phrases by predicting their tokens}.
Additionally, we propose an \ti{completeness regularization term} 
(the second term)
over the masked phrases to encourage complete phrase reconstruction,
i.e.,
the model will get more rewards when an \ti{entire phrase} is correctly predicted.
As defined in Eq.(5), 
where $\hat{\mc{P}^t}$ also denotes the indices of masked tokens but grouped by phrases,
 $P$ denotes one of the group in $\hat{\mc{P}^t}$,
we first average all the token embeddings of a phrase to obtain the merged phrase feature (i.e., $\tb{Avg}\big(\big[\mc{F}\big(X_{\backslash M};\theta\big)\big]_{P}\big)$ ).
Then we predict each complete phrase instead of the tokens in it using its merged feature along with a new phrase softmax matrix (i.e., $\bs{C^\top}$).
$\mc{V}_P$ represents the set of all phrases in corpus.
\begin{figure*}[t]
\setlength{\belowcaptionskip}{-6pt}
\centering
\includegraphics[width=0.9\textwidth]{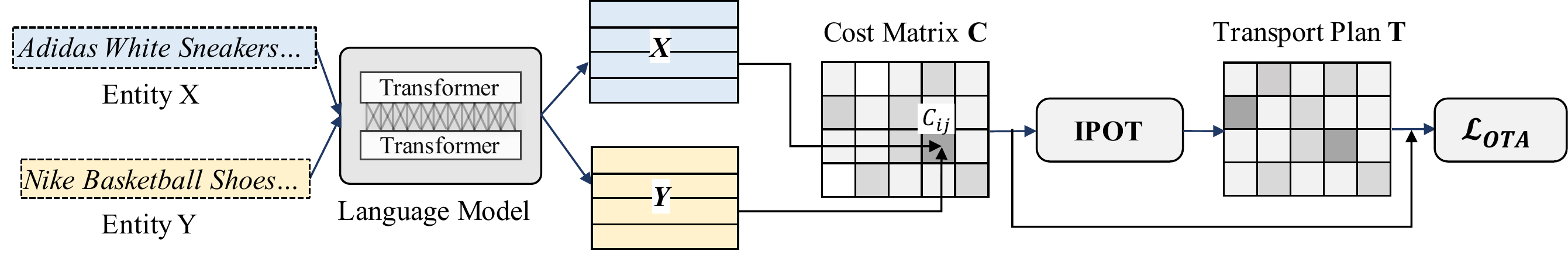}
\vspace{-0.2cm}
\caption{Illustration of the OT based approach for learning the word-level alignments
for entity association based text pair.
}
\label{ota}
\end{figure*}

\begin{savenotes}
\begin{algorithm}
\begin{algorithmic}[1]
\caption{Token sampling algorithm for the phrase mode.}
\label{al:dp}
\REQUIRE An sequence $X^t$; The domain phrase pool $\mc{P}_{D}$. 
\ENSURE Token indices of domain phrases, denoted by $\mc{P}^{t}$; Token indices grouped by phrases, denoted by $\hat{\mc{P}}^{t}$.
 \STATE Detect phrases\footnote{Fulfill via a rule-based phrase matcher,  https://spacy.io/usage/rule-based-matching} in $X_t$ that intersect with $\mc{P}_{D}$, denoted by $\mc{P}_{T}$;\\
 \STATE Retrieve their quality scores $\{s_i\}$ from $\mc{P}_{D}$;\\
 \STATE Normalize all the scores by softmax, i.e.,\\ $s_{n,i}=\exp(s_i)/\exp(\sum_j\exp(s_j))$;\\
 \STATE Let count $=0$, $\mc{P}^{t}=\emptyset$, $\hat{\mc{P}}^{t}=\emptyset$;\\
 \WHILE{count/\text{num\_token}($X_t$)$<$15\%}
 \STATE Sample a phrase $p$ from $\mc{P}_{T}$ based on the normalized \\scores, i.e., $\{s_{n,i}\}$;\\
 \STATE Add the indices of all tokens in $p$ into $\mc{P}^t$;\\
  \STATE Add the indices in $p$ as a list into $\hat{\mc{P}}^t$;\\
 \STATE count $+\!\!=1$;\\
 \ENDWHILE
  \STATE Return $\mc{P}^t$, $\hat{\mc{P}}^t$.
  \end{algorithmic}
\end{algorithm}
\end{savenotes}

\subsubsection{\tb{Adaptive Hybrid Learning}}
As both word-level and phrase-level semantics are critical to language modeling,
we combine the two learning modes via a dynamic parameter $\alpha$ based on the feedback losses of them.
At each iteration, as shown in Figure \ref{ahm}, the model automatically selects the weaker mode according to the value of $\alpha$.

\noindent{\tb{Calculating ${\bs{\alpha}}$}.} 
We calculate $\alpha$ based on the relative loss reduction speed of the two modes.
Specifically, at each iteration (assuming $t^{th}$), we first calculate a special variable for both modes to track their fitting progress, i.e., $\eta_{w}^{t}$ and $\eta_{p}^{t}$.
The larger $\eta_{w}^{t}$ ($\eta_{p}^{t}$) is, the less sufficient the model is trained on the word (phrase) mode.
Then $\alpha^{t+1}$ for next iteration is calculated as the rescaled ratio of $\eta_{w}^{t}$ and $\eta_{p}^{t}$, i.e.,
\begin{equation}
\small
\eta_{w}^{t}=\frac{\Delta^{t,t-1}_{w}}{\Delta^{t,1}_{w}}=\frac{\big[\mc{L}_{w}^{t-1}-\mc{L}_{w}^t\big]_{+}}{\mc{L}_{w}^{1}-\mc{L}_{w}^t}, \quad
\eta_{p}^{t}=\frac{\Delta^{t,t-1}_{p}}{\Delta^{t,1}_{p}}=\frac{\big[\mc{L}^{t-1}_{p}-\mc{L}^t_{p}\big]_{+}}{\mc{L}^{1}_{p}-\mc{L}^t_{p}},
\end{equation}
\begin{equation}
\small
 \alpha^{t+1}=\text{tanh}\big(\eta_{w}^{t+1}/\eta_{p}^{t+1}\big).
\end{equation}
where $\mc{L}_{w}^t$ denotes the loss of the word learning mode and will only be updated if word mode is selected at the $t$-th iteration.
Function $[x]_+$ is equivalent to $max(x,0)$.
$\Delta^{t,t-1}_{w}$ denotes the loss reduction of word mode between the current and last iteration.  
$\Delta^{t,1}_{w}$ denotes the total loss reduction.
$\Delta^{t,t-1}_{p}$, $\Delta^{t,1}_{p}$, $\mc{L}_{p}^t$ represents the same variables in the phrase mode.
Thus, $\eta_{w}^{t+1}$ and $\eta_{p}^{t+1}$ indicates the relative loss reduction speed of the two modes respectively,
and the ratio them ($\eta_{w}^{t+1}/\eta_{p}^{t+1}$) reflects the relative importance of the word mode.
The non-linear function $\text{tanh}$ is used to rescale the ratio to [0,1].

\noindent{\tb{Loss Function of \sahm}}.
The overall loss function of \sahm~is the combined losses of the two learning modes, with weights dynamically adjusted by $\alpha^t$, i.e.,
\begin{equation}
\begin{aligned}
\small
\mc{L}_{\text{AHM}} = \small\frac{1}{|\mc{D}|}\sum_{X^{t}\in\mc{D}}\mathbb{I}
(\alpha^{t})\cdot\mc{L}_{w}+\mathbb{I}(1-\alpha^{t})\cdot\mc{L}_p,
\end{aligned}
\end{equation}
\begin{equation}
\label{indicator}
\mathbb{I}(x) =
\begin{cases}
1 ~&\text{ if }~ x > 0.5, \\
0 ~&\text{ if }~ x \leq 0.5.
\end{cases}
\end{equation}
where $\mc{D}$ represents the training corpus. 
$\mathbb{I}$ denotes the indicator function defined in Eq.(\ref{indicator}).
As can be seen, when $\eta_{w}^{t+1}\gg\eta_{p}^{t+1}$, $ \alpha^{t+1}\approx1$, 
the word mode becomes dominating, and vice versa.
In other words, $\alpha$ is able to control the model to switch to the weaker learning mode adaptively.

\subsection{OT-based \np}

To exploit the co-occurrence signals hidden in entity associations, we formulate a new pre-training task, i.e., \np~(\snp), as defined below. 
We first exploit the entity association graph to extract a collection of \ti{associated text pairs} from the domain corpus as training data.
Next, an Optimal Transport (OT) based approach is introduced to train \snp~effectively.

\begin{definition}
	\label{def:CEA}
    Given two paired entity contents denoted by word sequences $\{{x_i}\}_{i=1}^m$ and $\{{y_j}\}_{j=1}^n$,
     \np~aims to learn an word-level alignment matrix $\tb{A}$, where $\tb{A}_{i,j}\in[0,1]$ indicates the correlation of $x_i$ and $y_j$ (\textrm{s.t.} $\sum_{j}\tb{A}_{i,j}=1,\sum_{i}\tb{A}_{i,j}=1$).
\end{definition}
     
The task is challenging due to the lack of groundtruth alignment matrix $\tb{A}'$.
A common solution to this problem involves designing advanced attention mechanisms to simulate soft alignment.
However, 
the learned attention matrices are often too dense and lack interpretability, inducing less effective alignment learning.
On the other hand, OT possesses \ti{ideal sparsity} that makes it a good choice for cross-domain alignment problems \cite{chen2020graph}. 
Specifically,
when solved exactly, 
OT yields a sparse solution $\tb{T}^{*}\in\sR^{m\times n}$ containing $(2r-1)$ non-zero elements at most, where $r = max(m, n)$, leading to a more interpretable and robust alignment.
Hence, we propose an OT-based approach to the address \snp.
Figure \ref{ota} presents an overview illustration of our Optimal Transport based approach for CEA.
Concretely, we follow the below procedures to fulfill it.

\noindent{\tb{Content Embeddings and Cost Matrix.}}
Given the entity pair $(X,Y)$, we first feed their content texts into the language model (Transformer) respectively to get the contextual embeddings, denoted by $\bs{X}=\{\bs{x_i}\}_{i=1}^m$ and $\bs{Y}=\{\bs{y_j}\}_{j=1}^n$.
Then we calculate a cost matrix $\tb{C}\in \sR^{m\times n}$, where $\tb{C}_{ij}$ defines the cost (distance) of shifting one mass from $\bs{x_i}$ to $\bs{y}_j$, where we use cosine distance $c(\xv_i,\yv_j)=1-\frac{\xv_i^\top\yv_j}{||\xv_i||_2 ||\yv_j||_2}$ as the cost function.

\noindent{\tb{Computing Transport Plan as Alignments}}.
Next, by regarding the two set of content embeddings $\bs{X},\bs{Y}$ as two probability distributions,
we calculate the optimal transport plan $\tb{T}^{*}\in \sR^{m\times n}$ of transforming one distribution to the other.
Here $\tb{T}^{*}$ is obtained via substituting $\bs{X},\bs{Y}$ into Eq.(2), i.e.,
\begin{equation}
\tb{T}^* = \argmin_{\Tmat\in \Gamma (\rvu,\rvv)}\left \langle \tb{T}, \tb{C} \right \rangle=\argmin_{\Tmat\in \Gamma (\rvu,\rvv)}\sum_{i=1}^m \sum_{j=1}^n \Tmat_{ij} \cdot c(\xv_i,\yv_j)\,,
\end{equation}
where each element $\tb{T}^* _{ij}$ in $\tb{T}^*$ denotes how much mass should be shifted from $\bs{x_i}$ to $\bs{y_j}$.
To be noted, the value of $\tb{T}^*_{ij}$ can be automatically optimized smaller if $\bs{x_i}$ and $\bs{y_j}$ are not very correlated, i.e., having a high cost value $\tb{C}_{ij}$.
In other words, $\tb{T}^*$ actually reflect the strength of correlations between the word-level content pair across two products.
Therefore, 
after jointly optimized with the language model,
we use $\tb{T}^*$ as the approximation to the alignment matrix. 

\noindent{\tb{Efficient Solver: IPOT}}.
Unfortunately, it is computational intractable \cite{arjovsky2017wasserstein, salimans2018improving} to compute the exact minimization over $\Tmat$.
Hence, to ensure an efficient training on large neural networks of language models,
we propose to apply the recent introduced Inexact Proximal point method for Optimal Transport (IPOT) algorithm \cite{xie2020fast} to compute the optimal transport plan $\tb{T}^*$.
IPOT approximates the exact solution by iteratively solving the following optimization problem:
\begin{equation}
\tb{T}^{(t+1)} = \argmin_{\Tmat\in \Gamma (\rvu,\rvv)}\Big\{\left \langle \tb{T}, \tb{C} \right \rangle +\beta\cdot\mc{B}(\tb{T},\tb{T}^{(t)}) \Big\}
\end{equation}
where $\mc{B}(\tb{T},\tb{T}^{(t)})$ is the proximity metric term used to penalizes solutions that are too distant from the latest approximation.
We do not choose Sinkhorn algorithm \cite{cuturi2013sinkhorn} to solve the efficiency issue as it is too sensitive to the choice of the hyper-parameter $\varepsilon$ in experiments.

\noindent{\tb{Loss Function of \snp}}.
Lastly, we train the language model via optimizing the OT distance (i.e., Wasserstein distance) between the aligned content embeddings, with overall loss function defined as:
\begin{equation}
\mc{L}_{OTA}(\bs{X,Y}) = \left \langle \tb{T}^*, \tb{C} \right \rangle=\sum_{i=1}^m \sum_{j=1}^n \Tmat^*_{ij} \cdot c(\xv_i,\yv_j)\,,
\end{equation}
\begin{equation}
\mc{L}_{OTA} = \frac{1}{|\mc{R}|}\sum_{(\bs{X,Y})\in\mc{R}} \mc{L}_{OTA}(\bs{X,Y})
\end{equation}
where $\mc{R}$ denotes the set of entity association based text pairs.



\section{Experiments}
In this section, we conduct extensive experiments in the e-commerce domain to validate the effectiveness of the proposed framework.
We first introduce the external and internal baselines compared in the paper.
Next, we present the corpus as well as the auxiliary domain knowledge data used during pre-training.
Besides, we elaborate the downstream tasks (definitions, datasets, performance metrics) for evaluating all the models.
Lastly, we report the main performance comparison, ablation studies, case studies, and visualization of the OT-based alignments. 


\subsection{Baseline Models}
\tb{External Baselines.} In this paper, we compare our framework to following external baselines.
	 (1) \tb{BERT:}~The vanilla BERT which is pre-trained on large-scale open-domain corpora by huggingface.
	(2) \tb{BERT-PT}~\cite{xu2019bert}:~The vanilla BERT that is further \textbf{p}ost-\textbf{t}rained on review data. This can be considered as the domain-oriented vanilla BERT.
	(3) \tb{BERT-NP:}~The vanilla BERT using a different masking strategy, i.e., masks \textbf{n}oun \textbf{p}hrases instead of words. We contrast this method with another internal baseline (\tb{DPM}) 
	to reveal the effects of different phrase selection schemes.
	(4) \tb{SpanBERT} \cite{joshi2020spanbert}: An variant of BERT which masks spans of tokens instead of individual tokens. We compare with it to further validate the effect of different masking schemes.
	(5) \tb{RoBERTa} \cite{liu2019roberta}: A robustly optimized variant of BERT which deletes the Next Sentence Prediction task.
	(6) \tb{ALBERT} \cite{lan2019albert}: A memory-efficient lite BERT that also high performances.
To enable the above baselines (2)-(6) to be \tb{domain-oriented}, like most existing work, we pre-train them on the same domain corpus as our method (except \tb{BERT} for validating the effects of using domain corpus).

\noindent\tb{Internal Baselines.}
For ablation studies (validating the effects of each component in framework), we further compare with the following internal baselines:
	(1) \tb{DPM:}~The vanilla BERT that only masks \textbf{d}omain \textbf{p}hrases using our phrase pool, abandons word masking. 
	(2) \tb{DPM-R:}~The vanilla BERT~that only masks \textbf{d}omain \textbf{p}hrases and further employs the phrase \tb{r}egularization term, abandons word masking.
	(3) \tb{HM-R:}~The vanilla BERT~that \tb{m}asks \textbf{d}omain \textbf{p}hrases and words in a \tb{h}ybrid way (50\%/50\% of the time), employs the phrase \tb{r}egularization term.
	(4) \tb{\sahm:} \ahm, without leveraging entity association knowledge by \np.
	(5) \tb{\sahm+\snp:}~The full version of our framework, combines \sahm, OT-based \snp~via continual multi-task learning.
    All internal baselines are pre-trained on the same domain corpus.

\subsection{Domain-oriented Tasks and Metrics}
We perform evaluations on four tasks of e-commerce. 
The definition, fine-tuning head and metric of each task is provided below.

\noindent{\tb{Review Question Answering (Review QA)}.}
Given a question $q = \{q_i\}_{i=1}^m$ about a product and a related review snippet $r = \{r_i\}_{i=1}^n$, it aims to find the span $s = \{r_i\}_{i=s}^e$ from $r$ that can answer $q$.
We employ the same BERT fine-tuning head \cite{devlin2019bert} as which on span-based QA to fine-tune this task, which maximizes the log-likelihoods of the correct start and end positions of the answer. 

\noindent{\tb{Review Aspect Extraction (Review AE)}.}
Given a review $r = \{r_i\}_{i=1}^n$, the task aims to find product aspects that reviewers have expressed opinions on.
It is typically formalized as a sequence labeling task \cite{xu2019bert}, in which each token is classified as one of $\{B,I,O\}$, and tokens between $B$ and $I$ are considered as extracted aspects.
Following \cite{xu2019bert}, we apply a dense layer and softmax layer on top of BERT output embeddings to predict the sequence labels.

\noindent{\tb{Review Aspect Sentiment Classification (Review ASC)}.}
Given an aspect $a = \{a_i\}_{i=1}^l$ and the review sentence $r = \{r_i\}_{i=1}^n$ where $a$ extracted from, this task aims to classify the sentiment polarity (positive, negative, or neutral) expressed on aspect $a$.
For fine-tuning, following \cite{xu2019bert}, both $a$ and $r$ are input into our framework, and we feed the \texttt{[CLS]} token to a dense layer and softmax layer to predict the polarity.
Training loss is the cross entropy on the polarities.


\noindent{\tb{Product Title Categorization (PTC)}}.
Given a product title $x = \{x_i\}_{i=1}^n$, the task aims to classify $x$ using a predefined category collection $\mc{C}$.
Each title may belong to multiple categories, hence being a multi-label classification problem.
We feed the embedding of \texttt{[CLS]} token to a dense layer and the multi-label classification head for fine-tuning.



\noindent{\tb{Evaluation Metrics}}.
For review QA, we adopt the standard evaluation script from SQuAD 1.1 \cite{rajpurkar2016squad} to report Precision, Recall, F\textsubscript{1} scores, and Exact Match (EM).
To evaluate review AE, we report Precision, Recall, and F\textsubscript{1} score.
For review ASC, we report Macro-F\textsubscript{1} and Accuracy following \cite{xu2019bert}.
Lastly, we adopt Accuracy (Acc), 
and Macro-F\textsubscript{1} 
to evaluate product title categorization. 

\subsection{Experimental Datasets}
\subsubsection{\tb{Pre-training Resources}}
In the paper, we collect and leverage a domain corpus and two domain knowledge datasets.
Table \ref{st} shows the datasets statistics and
below presents the detailed collecting steps.

\noindent{\tb{Domain Corpus}.}
We extract millions of product titles, descriptions, and reviews from the Amazon Dataset\cite{ni2019justifying} to build this corpus.
The entire corpus consists of two sub-corpus, i.e., product corpus and review corpus.
In the first corpus, each line corresponds to a product title and its description, while in the second, each line corresponds to a user comment on a specific product.
The corpus serves as the foundation for language models to learn essential semantics of the e-commerce domain.

\noindent{\tb{Domain Phrase Pool}.}
To build the \ec~domain phrase pool,
we extract one million domain phrases from the above corpus leveraging AutoPhrase\footnote{https://github.com/shangjingbo1226/AutoPhrase},
a high efficient phrase mining algorithm, 
which is able to generate a quality score for each phrase based on corpus-level statistics such as \ti{popularity, concordance, informativeness,} and \ti{completeness}. 
Moreover, we filter out phrases that have a score lower than 0.5 to keep \ti{quality domain phrases}.
Table \ref{phrase} shows the top-ranked phrases from six product categories.
\begin{table}[t]
\centering
\caption{The statistics of the pre-training datasets.}
\vspace{-0.3cm}
\includegraphics[width=0.46\textwidth]{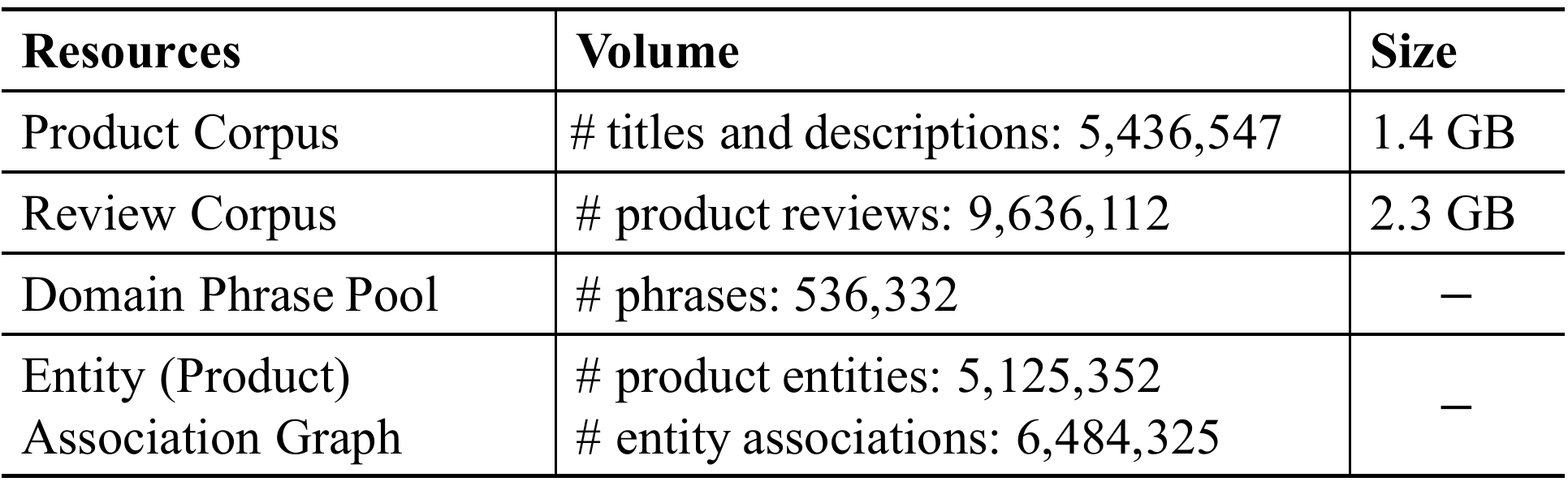}
\label{st}
\vspace{-0.3cm}
\end{table}

\begin{table}[t]
\renewcommand\arraystretch{1}
\small
  \centering
\caption{High-quality phrases of the e-commerce domain.}
\vspace{-0.3cm}
  \centering
  \begin{tabular}{m{2.3cm}<{\centering}|m{5.25cm}}
    \hline
    \tb{Category} & \tb{\quad \quad \quad Representative phrases} \\
    \hline
    \hline
     Automotive& jumper cables, cometic gasket, angel eyes, drink holder, static cling  \\
    \hline
     Clothing, Shoes and Jewelry&  high waisted jean, nike classic, removable tie, elegant victorian, vintage grey\\
     \hline
    Electronics & ipads tablets, SDHC memory card, memory bandwidth, auto switching \\
    \hline
    Office Products & decorative paper, heavy duty rubber, mailing labels, hybrid notebinder \\
    \hline
    Sports and Outdoors & basketball backboard, table tennis paddle, string oscillation, fishing tackles \\
    \hline
    Toys and Games & hulk hogan, augmented reality, teacup piggies, beam sabers, naruto uzumaki \\
    \hline
  \end{tabular}
\label{phrase}
\vspace{-0.3cm}
\end{table}

\noindent{\tb{Entity Association Graph}.}
We build this graph to store the product entity associations in the form of associated entity pairs.
In the paper, we only consider the ``substitutable'' associations and use a shopping pattern based heuristic method \cite{mcauley2015inferring} to extract corresponding product pairs with this relation.
We exploit all entity pairs in this graph to extract the same amount of associated text (title, description) pairs from the product corpus for the task of \snp.

\begin{figure}[t]
\centering
\includegraphics[width=0.48\textwidth]{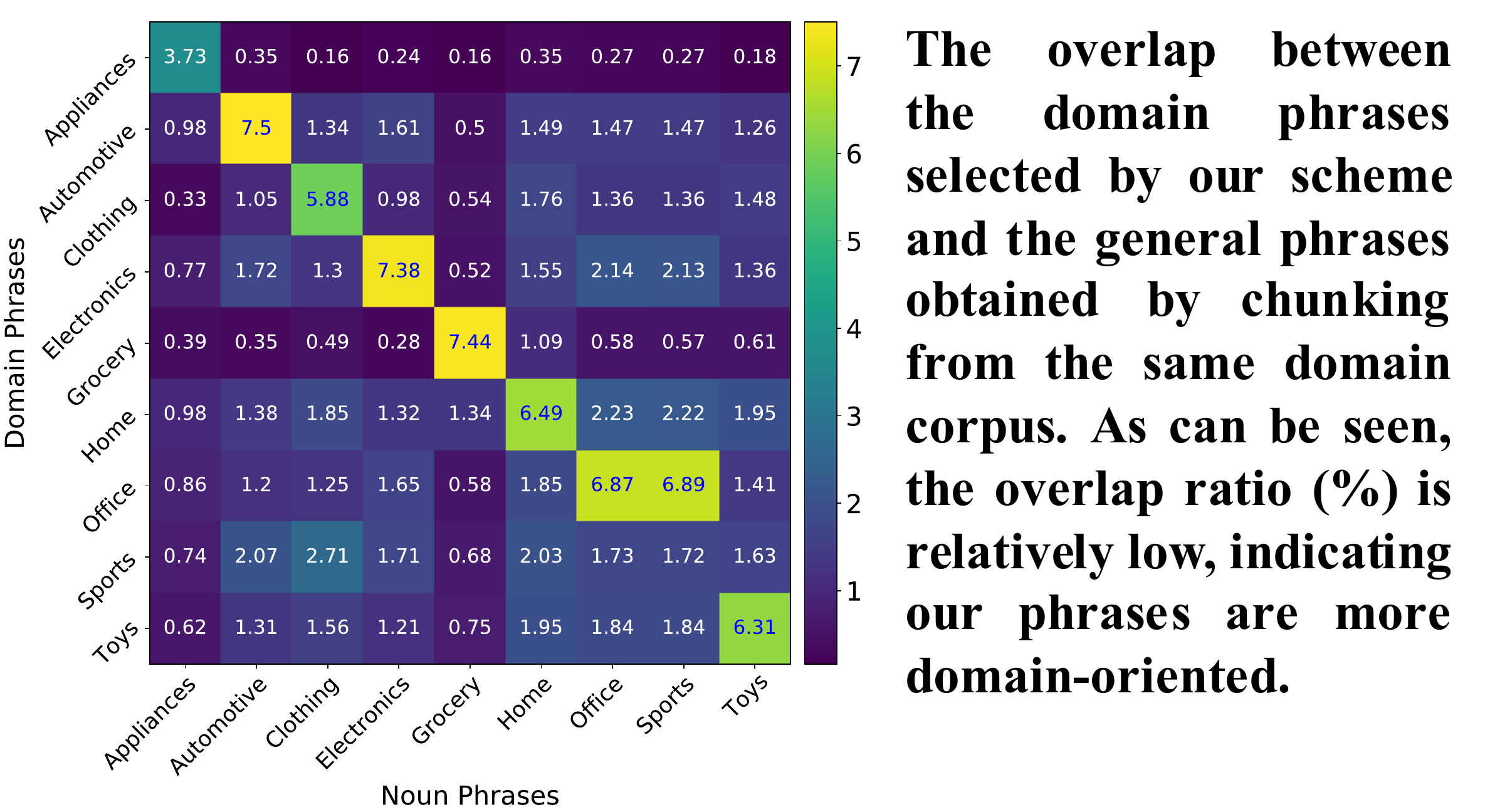}
\vspace{-0.5cm}
\caption{The overlap of different phrase sampling schemes.}
\label{overlap}
\vspace{-0.5cm}
\end{figure}

Figure \ref{overlap} presents when sampling phrases on the same domain corpus, the overlap between the results by our phrase pool based scheme and the ones by chunking based scheme. 
Results are reported based on nine categories of the product corpus.
Each entry represents the ratio of the overlapped phrases to the general chunking based phrases. 
As can be seen, the overlap ratio is relatively low across all the sub categories, indicating our phrase pool based scheme yields more domain-oriented phrases.





\subsubsection{\tb{Task-specific Datasets}}
For review QA, we evaluate on a newly released Amazon QA dataset \cite{miller2020effect},  consisting of 8,967 product-related QA pairs.
For the task of review AE and review ASC, we employ the laptop dataset of SemEval 2014 Task 4 \cite{pontiki2016semeval}
which contains 3,845 review sentences, 3,012 annotated aspects and the sentiment polarities on them.
For product title categorization, we create an evaluation dataset by extracting a subset of Amazon metadata, consisting of 10,039 product titles and 98 categories.
The first three datasets above are publicly available from prior works and we will share the fourth dataset in future.
For all the datasets, we divide them into training/validation/testing set with the ratio of 7:1:2.

\begin{table*}[!t]
\centering
\caption{Performance comparison of baselines and our model on the e-commerce downstream tasks (\%).}
\vspace{-0.2cm}
\includegraphics[width=0.97\textwidth]{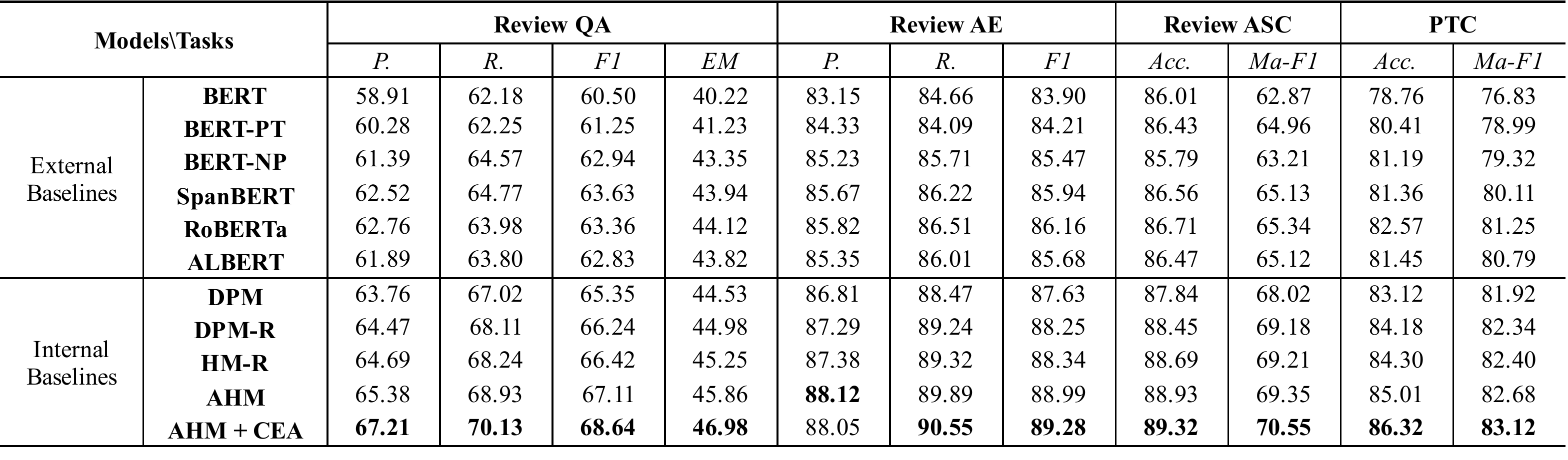}
\label{result}
\vspace{-0.2cm}
\end{table*}

\subsection{Implementation Details}

\noindent{\tb{Pre-training details}.} 
All the models are initialized
with the same pre-trained BERT (the \texttt{bert-base-uncased} by Huggingface, with 12 layers, 768 hidden dimensions, 12 heads, 110M parameters). 
We post-train all the models (except BERT) on the domain corpus for 20 epochs, with batch size 32 and learning rate 1e-5.
For our framework, we adopt Continual Multi-task Learning \cite{sun2020ernie} to combine \sahm~and \snp.
Specifically, we first train \sahm~alone on the entire corpus for $10$ epochs with the same batch size and learning rate. 
Then, we train \sahm~and \snp~jointly on the product corpus (with instances reformatted as text pairs by entity associations) for another $10$ epochs.
In \sahm, to initialize $\alpha$ and ensure a stable training, we fix $\alpha^t=0.6$ for t=1$\sim$1,000 (word learning mode is easier and provides preliminary knowledge, hence we weigh it more for initial iterations).
For training OT-based \snp, we set $\beta=0.5$ in the IPOT algorithm.
All the pre-training is performed on a computational cluster with 8 NVIDIA GTX-1080-Ti GPUs with 20 days duration.

\noindent{\tb{Fine-tuning Details.}} 
In each task, we adopt the same task-specific architecture (task head) as aforementioned for all the models.
We choose the learning rate and epochs from \{5e-6, 1e-5, 2e-5, 5e-5\} and \{2,3,4,5\} respectively.
For each task and each model, we pick the best learning rate and number of epochs on the development set and report the corresponding test results.
We found the setting that works best across most tasks and models is 2 or 4 epochs and a learning rate of 2e-5.
Results are reported as averages of 10 runs.

\subsection{Experimental Results}
\subsubsection{\tb{Main Results Analysis}}
Table \ref{result} presents the performance comparison of all the baselines and our framework on the four tasks.
The key observations and conclusions are:
(1) Our framework (\tb{AHM, AHM+CEA}) easily outperforms all the external baselines by a large margin (\tb{4.1\%} in average), 
indicating the effectiveness of our general idea, i.e., leveraging auxiliary domain knowledge to enhance domain-oriented language modeling.
(2) \tb{BERT-PT} outperforms \tb{BERT}, proving that for domain-oriented tasks, capturing domain semantics by pre-training on a domain corpus is necessary.
(3) \ti{The effects of different masking schemes}: \tb{BERT-NP} and \tb{SpanBERT} can perform better consistently than \tb{BERT-PT}, indicating the advantage of phrase/span based masking strategy over word based masking strategy.
(4) \ti{The effects of different phrase selection schemes}: \tb{DPM} achieves more improvements over \tb{BERT-NT} and \tb{SpanBERT}, certificating that the domain phrase pool based sampling outperforms general chunking based phrase sampling.
We attribute this to that: the domain phrase pool, 
serving as a ``supervisor'', 
enables the language model to ``focus'' more on \ti{domain-oriented phrases}, and these phrases have more effects over the downstream tasks.


\subsubsection{\tb{Ablation Studies}}
The bottom of Table \ref{result} shows the performance comparison of the internal baselines.
As can be seen,
(1) \tb{DPM-R} outperforms \tb{DPM}, validating the effectiveness of the proposed \ti{phrase regularization term}.
Compared with reconstructing phrases by tokens, it encourages complete phrase reconstruction, leads to a more accurate phrase perception learning.
(2) \tb{HM-R} uses hybrid masking in a straightforward way, achieves slightly better performances than \tb{DPM-R}.
Besides,
\tb{AHM} achieves more improvements on \tb{DPM-R} than \tb{HM-R}.
This indicates that both word learning and phrase learning are essential for language models, and the adaptive hybrid learning method is a more solid way to combine them.
(3) \tb{AHM+CEA} further improves the performances by \tb{0.5\%$\sim$1.2\%} over \tb{AHM} on the four tasks, 
certificating the effectiveness of our idea of leveraging entity association knowledge to augment semantic learning.
Moreover, the proposed OT-based alignment pre-train task can successfully exploit the hidden co-occurrence signals in entity association based text pairs.

\subsubsection{\tb{Case Studies and Visualizations}}
Table \ref{case} shows a case study of the review aspect extraction task.
We compare our model with {BERT-PT}, both are pre-trained on the same domain corpus, and employ the same fine-tuning architecture and task-specific dataset.
As can be seen, for ``aspects'' that span multiple words,
our model offers better predictions than BERT-PT in terms of the phrase completeness (\ti{size of the screen} vs \ti{screen}) .
This indicates that our model indeed possesses fine phrase perception ability needed for \ti{phrase-intensive tasks}.

\begin{table}[t]
\centering
\caption{
Case studies of Aspect Extraction (AE). Given a review, it aims to extract specific product ``aspects'' that are discussed. Ground-truth answers are marked in color.
For answers consisting of multi-word phrases, our model make more comprehensive predictions than BERT.
}
\includegraphics[width=0.47\textwidth]{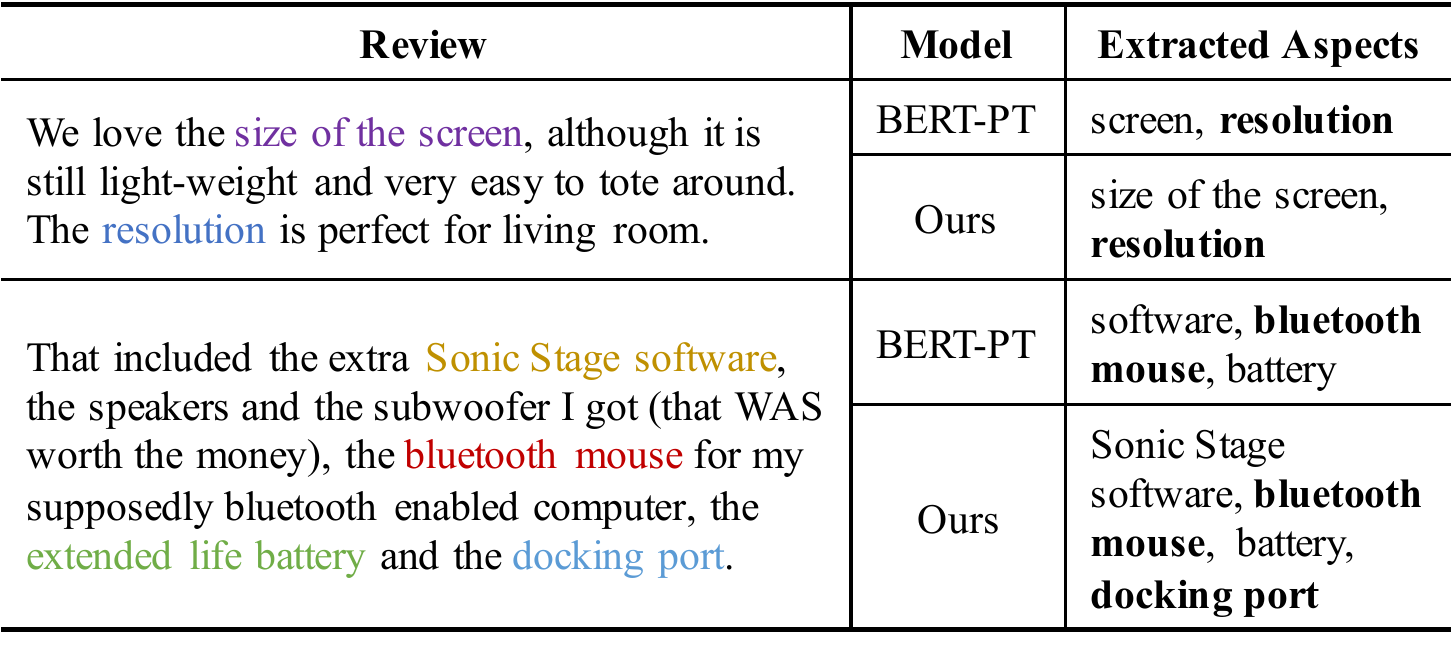}
\label{case}
\vspace{-0.4cm}
\end{table}

\begin{figure}[t]
	\centering
	\begin{tabular}{ @{}c@{ }c@{ }}
		\includegraphics[width=0.24\textwidth]{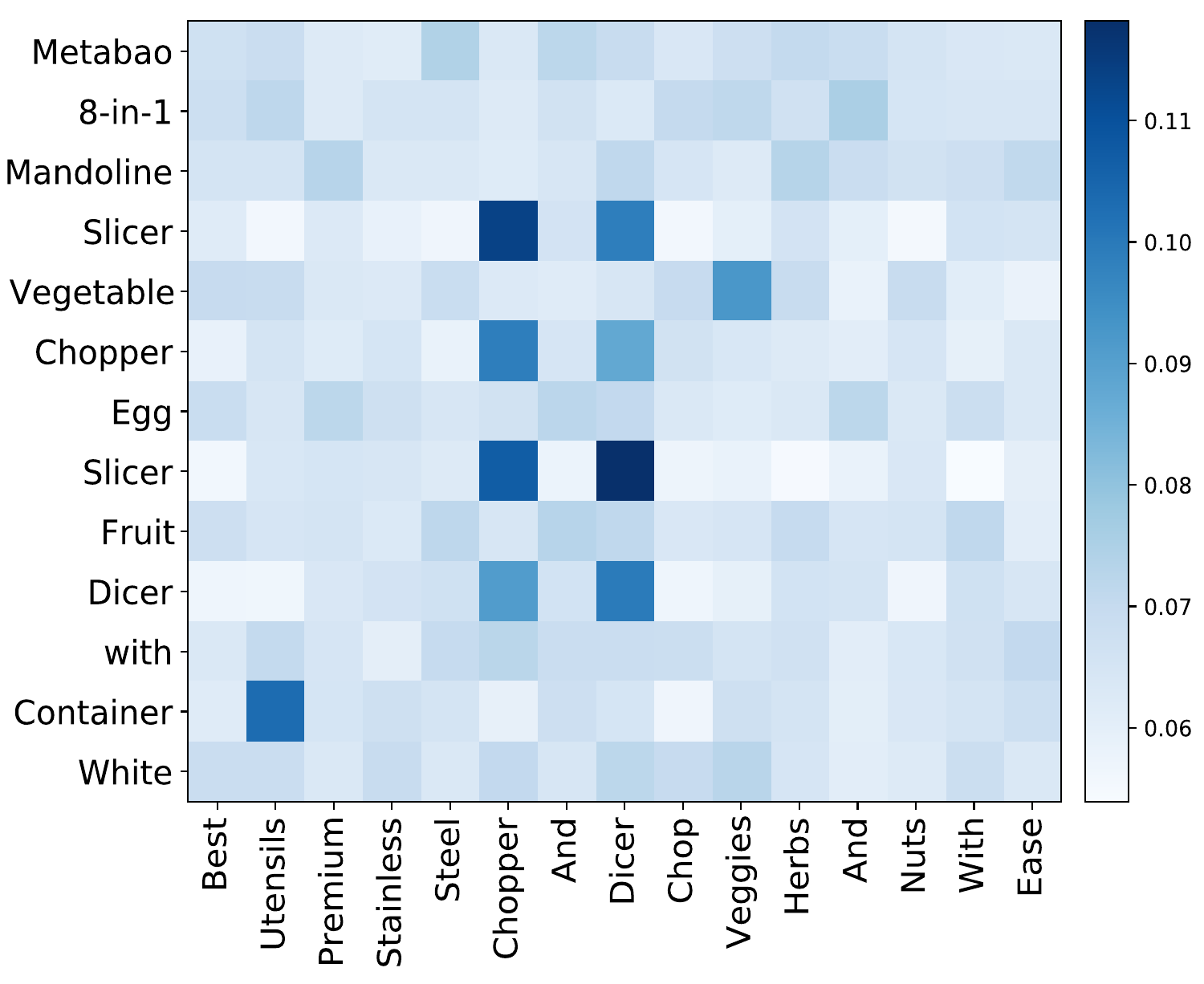}&
		\includegraphics[width=0.24\textwidth]{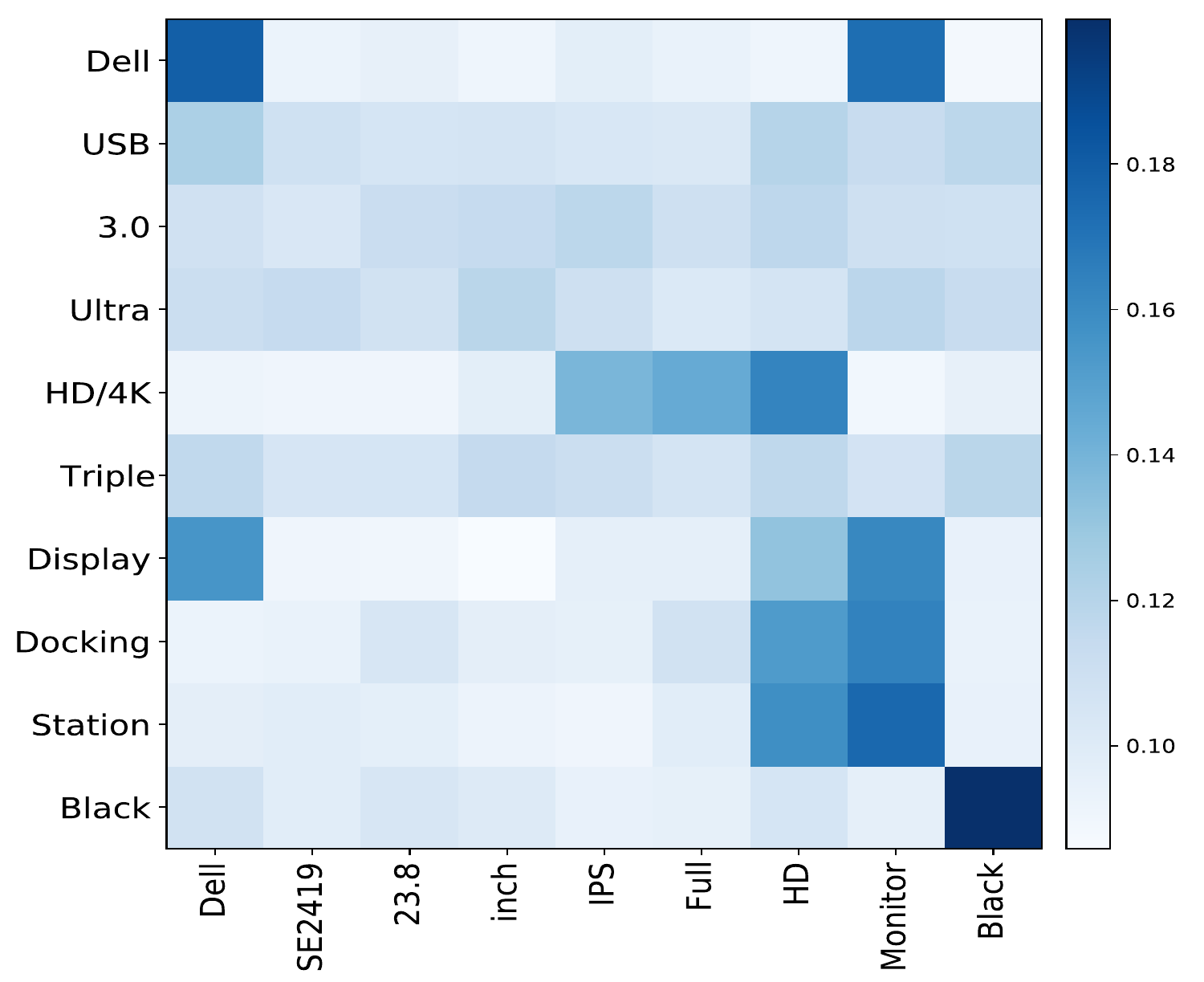}
		\\ 
	 (a) Two kitchen products. &
	 (b) Two electronic products.
	\end{tabular}
	\vspace{-0.2cm}
	\caption{Visualizing the optimal transport plan in two real examples. }
	\label{visual}
		\vspace{-0.4cm}
\end{figure}

Figure \ref{visual} presents the visualization of the optimal transport alignment for two product pairs, where darker color indicates stronger correlations.
Example (a) is about \texttt{Mandoline Slicer} and \texttt{Steel Chopper}, example (b)
is about a \texttt{Docking station} and \texttt{Dell Monitor}.
As can be seen, in both examples,
OT alignments are sparser and offers better interpretability, with meaningful word alignment pairs being discovered automatically
(\ti{Slicer} vs \ti{Chopper}, \ti{Vegetables} vs \ti{Veggies}, \ti{Monitor} vs \ti{Display}).
This certificates that the OT-based alignment task can indeed benefit semantic learning by automatically correlating similar words/phrases across entity pairs.

\section{Conclusion}
In this paper, we introduced how to improve domain-oriented language modeling by leveraging auxiliary domain knowledge. 
Specifically, we proposed a generalized pre-training framework enhancing existing works from two perspectives. 
First, we developed \ahm~(\sahm) to incorporate auxiliary domain phrase knowledge. 
Second, we designed \np~(\snp) to leverage entity association as weak supervision for augmenting the semantic learning of pre-trained models. 
Without the loss of generalization, we performed the experimental validation on four downstream e-commerce tasks. 
The results showed that incorporating phrase knowledge via \sahm~can improve the performance on all the tasks, especially the phrase-intensive ones. 
Also, utilizing the entity association knowledge via \snp~can further improve the performances and the learned alignments revealed meaningful semantic correlation across word pairs.

\bibliographystyle{ACM-Reference-Format}

\bibliography{ref}

\end{document}